# A polynomial algorithm for computing the optimal repair strategy in a system with independent component failures

**Sampath Srinivas**
Computer Science Department
Stanford University
Stanford, CA 94305
srinivas@cs.stanford.edu

## Abstract

The goal of diagnosis is to compute good repair strategies in response to anomalous system behavior. In a decision theoretic framework, a good repair strategy has low expected cost. In a general formulation of the problem, the computation of the optimal (lowest expected cost) repair strategy for a system with multiple faults is intractable. In this paper, we consider an interesting and natural restriction on the behavior of the system being diagnosed: (a) the system exhibits faulty behavior if and only if one or more components is malfunctioning. (b) The failures of the system components are independent. Given this restriction on system behavior, we develop a polynomial time algorithm for computing the optimal repair strategy. We then go on to introduce a system hierarchy and the notion of inspecting (testing) components before repair. We develop a linear time algorithm for computing an optimal repair strategy for the hierarchical system which includes both repair and inspection.

## 1 INTRODUCTION

The goal of doing diagnosis is to recommend good repair and maintenance actions in response to inferences about the state of the system. A repair *strategy* is a set of situation-action rules. The situations are the various possible observations and the actions are repair actions in response to these observations. In a decision theoretic framework, the goal is to compute repair strategies which have low expected cost. The investigation of methods for computing optimal (i.e., lowest cost) repair strategies is thus of great interest.

The computation of the optimal repair strategy in a general formulation of the repair problem is intractable (see for example, [Heckerman *et al*, 1995] for a discussion). In this paper, we set up an interesting restricted formulation of the repair problem in a system with multiple faults. The primary restriction is an assumption about how the system behaves – the system is assumed to exhibit faulty behavior if and only if one or more components have failed. Component failures are assumed to be independent. For example, in modeling a car, it may be reasonable to assume that the car runs normally iff both the fuel delivery subsystem and the ignition subsystem function normally. Further, one might assume that the failures of these subsystems are independent.

We analyze this restricted formulation of the repair problem and develop a polynomial time algorithm to compute the optimal repair strategy. At this point in the development, a strategy specifies a sequence in which the subsystems are to be repaired. Thus, the optimal strategy, for example, may specify that the fuel delivery system be repaired before the ignition system.

We then introduce a notion of hierarchy into our formulation and extend the polynomial algorithm mentioned above to compute optimal repair strategies for hierarchical systems. Extending the car example, one could model the fuel delivery subsystem as consisting of the fuel pump and carburetor. Say the fuel subsystem works iff both these subcomponents work normally. A hierarchical repair strategy specifies, as before, the sequence in which the fuel delivery and ignition subsystems should be repaired. Further, for each subsystem it specifies the order in which its subcomponents should be inspected and repaired. Thus the strategy may specify that the fuel pump be inspected and repaired before the carburetor.

This paper is structured as follows: In the next section, we describe our repair model in more detail. We then derive a condition for a repair strategy to be optimal in the general case where the component failures may be correlated. When the component failures are independent, we show that this optimality condition has a simple form. This simple form allows the optimal strategy to be computed in polynomial time with a simple sorting procedure.

Up to this point, the only kind of repair action that is allowed is the replacement of components. We now go on to introduce another class of repair actions, viz, component inspections. A component inspection tests whether a component is working or not. Following this, we generalize the notion of a system and introduce a system hierarchy. The optimality condition mentioned above is then used to develop a linear time algorithm for computing the best hierarchical repair strategy for a hierarchical system. A hierarchical repair strategy includes both inspection and component replacement actions. We conclude by examining related work.

## 2 DEFINING THE PROBLEM

Consider a system with $n$ components $C_i$, $1 \leq i \leq n$, for which we want to develop good repair strategies. Say each component can be either be in an ok state ($ok$) or broken state ($b$). The state of $C_i$ is represented by a mode variable $M_i$. Say we are given some joint distribution $P(M_1, M_2, \ldots, M_n)$ of the mode variables. In addition, we are given a repair cost $c_i$ for each component $C_i$. After a component is repaired we assume that it is in the $ok$ state. The cost $c_i$ can also be interpreted as the cost of replacing $C_i$.

We assume that the system works normally only if all the components are in the $ok$ state. If any of the components are in the $b$ state the system exhibits a fault. The system status $X$ is assumed to be observable. If $X = ok$ then it means the system is working normally. If $X = b$ it means the system is broken (i.e., exhibiting a fault).

Say the repair protocol is as follows – we will observe the system status $X_0$ before we choose any fix action. If $X_0 = ok$ we stop. If $X_0 = b$, then we choose to fix some component $C_1$ and then observe the system status $X_1$. If $X_1 = ok$ we stop. If $X_1 = b$ we continue, choosing some other component $C_2$ to fix and so on. We will refer to the action of fixing $C_j$ as $fix_j$. A repair *strategy* is a sequence in which to examine components in the repair protocol described above.

Consider a strategy $T = \langle C_1, C_2, C_3, \ldots, C_n \rangle$. Say that the first $k - 1$ components have been repaired according to strategy $T$ and the system is still faulty. We will refer to the sequence of observations and actions up to this point as $S_{k-1}$. Hence, $S_{k-1} = \langle X_0 = b, fix_1, X_1 = b, fix_2, X_2 = b, \ldots, fix_{k-1}, X_{k-1} = b \rangle$. After observing $S_{k-1}$ the next step is to replace $C_k$ at a cost of $c_k$. Further, if we replace $C_k$ and the system is still faulty we replace $C_{k+1}$ incurring cost $c_{k+1}$. This event occurs with a probability $P(fix_k, X_k = b|S_{k-1})$.

We can thus compute the expected cost $EC(T)$ of the strategy $T$ as:

$$\begin{aligned}
EC(T) =\ & P(S_0)[c_1 + \\
& P(fix_1, X_1 = b|S_0)[c_2 + \\
& P(fix_2, X_2 = b|S_1)[c_3 + \\
& \ldots \\
& P(fix_m, X_m = b|S_{m-1})[c_{m+1}
\end{aligned} \quad (1)$$

$$\ldots$$
$$+ P(fix_{n-1}, X_{n-1} = b|S_{n-2})c_n \ldots] \ldots]]]$$

We note from the definition of $S_k$ that $S_k = \langle S_{k-1}, fix_k, X_k = b \rangle$. Hence we have:

$$P(fix_k, X_k = b|S_{k-1})P(S_{k-1}) = $$
$$P(S_{k-1}, fix_k, X_k = b) = P(S_k)$$

Using this result repeatedly in Equation 1 we find that the expression telescopes to:

$$EC(T) = \Sigma_{1 \leq k \leq n}\ c_k \times P(S_{k-1}) \quad (2)$$

### 2.1 AN EXPRESSION FOR $P(S_k)$

Let a world be a state assignment to all the mode variables of the system. Consider the worlds which are *inconsistent* with the observation $S_k = \langle X_0 = b, fix_1, X_1 = b, fix_2, X_2 = b, \ldots, fix_k, X_k = b \rangle$. A world $w$ is inconsistent with $S_k$ iff the observation $S_k$ could *not* have occurred if the true state of the system was $w$.

We see that the worlds *inconsistent* with $S_k$ are exactly those worlds $\hat{w}$ in which *all* the components which have not yet been fixed are in the $ok$ state. The reason is as follows. If any of the worlds $\hat{w}$ had been the true situation, then we know that the broken components are some subset of $\{C_i | 1 \leq i \leq k\}$. Hence the repair sequence $\langle C_1, C_2, \ldots, C_k \rangle$ would necessarily have resulted in $X_j = ok$ for some $j \leq k$ (when all the broken components were fixed). Since such an observation is inconsistent with $S_k$, we conclude that $\hat{w}$ is inconsistent with $S_k$.

By a similar line of argument, we can conclude that any world in which at least one of the remaining unfixed components is broken is consistent with $S_k$. The total probability mass of the worlds in which all of $C_{k+1}$, $C_{k+2}$, $\ldots$, $C_n$ are in the $ok$ state is $P(M_{k+1} = ok, M_{k+2} = ok, \ldots, M_n = ok)$. Hence $P(S_k) = 1 - P(M_{k+1} = ok, M_{k+2} = ok, \ldots, M_n = ok)$. We will use the notation $M_{[i,j]} = ok$ as a short form for $\langle M_i = ok, M_{i+1} = ok, \ldots, M_j = ok \rangle$. Hence, Equation 2 simplifies to:

$$EC(T) = \Sigma_{1 \leq k \leq n}\ c_k \times [1 - P(M_{[k,n]} = ok)] \quad (3)$$

## 3 THE OPTIMALITY CONDITION

We will now derive a condition under which a strategy is optimal (i.e, has the lowest possible expected cost).

Consider a strategy $T^j = \langle C_1, C_2, \ldots C_j, C_{j+1}, \ldots, C_n \rangle$. Let $T^{j+1}$ be identical to $T^j$ except that the positions of the $C_j$ and $C_{j+1}$ are transposed in the sequence. We compare the expected costs of $T^j$ and $T^{j+1}$. We have:

$$EC(T^j) - EC(T^{j+1}) = \quad (4)$$
$$(c_j[1 - P(M_{[j,n]} = ok)] +$$

$$c_{j+1}[1 - P(M_{[j+1,n]} = ok)])$$
$$-(c_{j+1}[1 - P(M_{[j,n]} = ok)] +$$
$$c_j[1 - P(M_j = ok, M_{[j+2,n]} = ok)])$$

Strategy $T^j$ is less expensive that $T^{j+1}$ if $EC(T^j) - EC(T^{j+1}) \leq 0$. Let us use the notation $R_{ok}$ for the event $M_{[j+2,n]} = ok$. Simplifying Equation 4 the condition $EC(T^j) - EC(T^{j+1}) \leq 0$ simplifies to:

$$c_j[P(M_j = ok, R_{ok}) -$$
$$P(M_j = ok, M_{j+1} = ok, R_{ok})]$$
$$\leq c_{j+1}[P(M_{j+1} = ok, R_{ok}) -$$
$$P(M_{j+1} = ok, M_j = ok, R_{ok})] \quad (5)$$

Thus, given a distribution $P(M_1, M_2, \ldots, M_n)$ and a strategy $T$, we can check whether the strategy is a (local) optimum by checking whether Equation 5 holds for adjacent components in the strategy. If the condition does hold for every pair of adjacent components the strategy is a local optimum. That is, exchanging the order of any two adjacent components in the strategy will always lead to a strategy with increased cost. Note that computation of the probabilities needed in Equation 5 from the joint distribution $P(M_1, M_2, \ldots, M_n)$ can be expensive. We will see, however, that the optimality condition takes a simple form when the failures of the components are independent.

### 3.1  A SANITY CHECK: THE SINGLE FAULT CASE

We have derived the above condition assuming a general distribution $P(M_1, M_2, \ldots, M_n)$. We now show that if we enforce a single fault assumption, Equation 5 reduces to the optimality condition of [Kalagnanam & Henrion, 1988] (see Section 7). They prove that in the case of a single fault, the optimal strategy replaces components in increasing order of the ratio $\frac{c_i}{p_i}$ where $c_i$ is the cost of replacement of $C_i$ and $p_i$ is the prior probability that $C_i$ is faulty.

Consider a single fault distribution. There are only $n$ possible worlds. Let these worlds be $w_1, w_2, \ldots, w_n$. $w_i$ is the world in which $M_i$ is in the $b$ state and all the other $M_j$ (i.e., $j \neq i$) are in the $ok$ state. Let the probability of world $w_i$ be $p_i$. That is, the probability that $C_i$ is the (only) faulty component is $p_i$.

Consider the probability $P(M_j = ok, R_{ok})$ in Equation 5. The worlds consistent with $\langle M_j = ok, R_{ok}\rangle$ are $w_1, w_2, \ldots, w_{j-1}$ and $w_{j+1}$. Hence $P(M_j = ok, R_{ok}) = (\Sigma_{[1 \leq i \leq j-1]} p_i) + p_{j+1}$. Call the quantity $(\Sigma_{[1 \leq i \leq j-1]} p_i)$ as $p_{prev}$. We have:

$$P(M_j = ok, R_{ok}) = p_{prev} + p_{j+1}$$

Using a similar line of reasoning:

$$P(M_j = ok, M_{j+1} = ok, R_{ok}) = p_{prev}$$

Hence the first term of Equation 5 reduces to $c_j[(p_{prev} + p_{j+1}) - p_{prev}] = c_j p_{j+1}$. Symmetrically, the second term reduces to $c_{j+1} p_j$. This simplifies Equation 5 to the result in [Kalagnanam & Henrion, 1988]:

$$\frac{c_j}{p_j} \leq \frac{c_{j+1}}{p_{j+1}}$$

In this special case, a strategy which satisfies the condition for every pair of adjacent components is globally optimal.

## 4  INDEPENDENT FAULTS

Say that each component $C_i$ can fail independently with probability $p_i$. That is, $P(M_i = b) = p_i$. We derive a simplification of the optimality condition (Equation 5) for this case.

Consider the first term of Equation 5. In this special case of multiple independent faults we have $P(M_j = ok, R_{ok}) = P(M_j = ok)P(R_{ok}) = (1 - p_j)P(R_{ok})$. Similarly $P(M_j = ok, M_{j+1} = ok, R_{ok}) = (1 - p_j)(1 - p_{j+1})P(R_{ok})$. The first term of Equation 5 hence becomes $P(R_{ok})c_j[(1 - p_j)p_{j+1}]$. Symmetrically, the second term of Equation 5 becomes $P(R_{ok})c_{j+1}[(1 - p_{j+1})p_j]$. Hence Equation 5 reduces to:

$$c_j \frac{1 - p_j}{p_j} \leq c_{j+1} \frac{1 - p_{j+1}}{p_{j+1}} \quad (6)$$

From this result we note that we can compute the globally optimal strategy by sorting the components $i$ by the quantity $(c_i \frac{1-p_i}{p_i})$. For an $n$ component system, this can be done in $O(n \log n)$.

We get an expression for the expected cost of any strategy (including the optimal strategy) by simplifying Equation 3 for the case of multiple independent faults. This gives:

$$EC(T) = \Sigma_{1 \leq k \leq n} \ c_k \times [1 - \Pi_{k \leq i \leq n}(1 - p_i)] \quad (7)$$

## 5  INTRODUCING COMPONENT INSPECTION

In the discussion so far, we have assumed that the only kind of action that is allowed is the replacement of a component. We now introduce the notion of inspecting a component. Inspection of a component $C_i$ determines what state it is in. Hence, if we inspect $C_i$ and find that it is in the $ok$ state, we do not have to take any further action to to fix the component. Carrying out the inspection of $C_i$ costs $d_i$. This cost is specified by the user. We note that $d_i \leq c_i$. If this was not the case there would no incentive to inspect the component – we could always replace it at less cost.

Now say that $C_i$ has been inspected and found to be broken. In that case, we will assume that it can be repaired incurring cost $H_i$. This cost is specified by the user. Note that $H_i \leq c_i$. If this was not the case, we would always replace the component rather than repair it incurring cost $H_i$.

Say we want to find an optimal strategy for repair under these conditions. A strategy specifies an order in which to repair the components (as before). In addition, for each component it also specifies whether the component is to be inspected before repair or not. An optimal strategy is the strategy with least expected cost.

Consider a strategy $T_{ins}^m = \langle\ [\mathcal{C}_1, rep], [\mathcal{C}_2, rep], \ldots, [\mathcal{C}_m, ins], \ldots, [\mathcal{C}_n, rep]\ \rangle$. The notation $ins$ says that the associated component is to be inspected. The notation $rep$ says that the associated component is to be simply replaced without inspection. Note that $T_{ins}^m$ specifies that all components except $\mathcal{C}_m$ be replaced without inspection. $\mathcal{C}_m$ alone is inspected before it is repaired. We now compute the expected cost of strategy $T_{ins}^m$.

The cost of $T_{ins}^m$ can be computed by simply replacing $c_m$ in Equation 1 by $d_m + P(M_m = b|S_{m-1})H_m$. That is, instead of the replacement cost $c_m$ we have to pay the inspection cost $d_m$. In addition, if component $m$ is indeed broken we have to pay cost $H_m$. The probability that we will find that $m$ is broken after inspection is $P(M_m = b|\Omega)$ where $\Omega$ is the current state of information. $\Omega$ includes all observations up to the replacement of $\mathcal{C}_{m-1}$ and the subsequent observation that the system is still not functioning (i.e., $X_{m-1} = b$). Hence $\Omega = S_{m-1}$.

Simplifying Equation 1 gives us:

$$\begin{aligned} EC(T_{ins}^m) &= [\Sigma_{1 \leq j < m}\ c_j \times P(S_{j-1})] \\ &\quad + d_m P(S_{m-1}) \\ &\quad + [\Sigma_{m < j \leq n}\ c_j \times P(S_{j-1})] \\ &\quad + H_m P(M_m = b, S_{m-1}) \end{aligned} \quad (8)$$

Consider the probability $P(M_m = b, S_{m-1})$. We saw before that the worlds inconsistent with $S_{m-1}$ are those worlds in which the remaining unrepaired components, $\mathcal{C}_m, \mathcal{C}_{m+1}, \ldots, \mathcal{C}_n$ are all in the $ok$ state. Note that $M_m = b$ is inconsistent with all of these worlds. Hence the set of worlds $consistent$ with $M_m = b$ (call the set $W_1$) is a subset of the set of worlds $consistent$ with $S_{m-1}$ (call this set $W_2$). As a result, we have:

$$\begin{aligned} P(M_m = b, S_{m-1}) &= P(W_1 \cap W_2) \\ &= P(W_1) \\ &= P(M_m = b) \end{aligned}$$

Hence the trailing term in Equation 8 reduces to $H_m P(M_m = b)$. Note that this term is not dependent on the position of $\mathcal{C}_m$ in the repair sequence.

In general, if we are given a strategy $T_P$ where some subset $P$ of the components are inspected we can come up with an expression for the cost as follows. Start with the expression for the case where you assume every element in the strategy $T_P$ is replaced without inspection (i.e., start with Equation 2). In this expression, replace $c_j$ by $d_j$ for each component $j$ which is in $P$, For each such component $j$, also add a trailing constant term $H_j P(M_j = b)$.

We now consider how we might compute an optimal strategy given a subset of components $P$ to be inspected. We will assume that the component failures are independent. We note that given any strategy $T_P$ which inspects just the components in $P$, the expression for the cost consists of two parts. One part is similar to Equation 2. The other part consists of constant terms of the type $H_j P(M_j = b)$ where $j \in P$. The latter part is unaffected by the order in which the components appear in strategy $T_P$.

Therefore to find the optimal strategy $T_P^{min}$ we have to only minimize the first part of the cost expression. Since the component failures are independent, we can directly apply Equation 6 to find the optimal sequence in which to repair the components. This sequence satisfies:

$$cd_j \frac{1 - p_j}{p_j} \leq cd_{j+1} \frac{1 - p_{j+1}}{p_{j+1}} \quad (9)$$

where:

$$cd_j = \begin{cases} d_j & \text{if } j \in P \\ c_j & \text{if } j \notin P \end{cases}$$

Given the optimal strategy $T_P^{min}$ the optimal repair cost can be computed as:

$$\begin{aligned} EC(T_P^{min}) &= \\ &\Sigma_{1 \leq k \leq n}\ cd_k \times [1 - \Pi_{k \leq i \leq n}(1 - p_i)] \\ &+ \Sigma_{\mathcal{C}_i \in P} H_i P(M_i = b) \end{aligned} \quad (10)$$

Thus, given a subset $P$ of components which are to be inspected, we can compute an optimal strategy and its cost in $O(n \log n)$.

### 5.1 THE GLOBALLY OPTIMAL STRATEGY

Say we consider every possible subset $P$ of the set of components and compute $T_P^{min}$. The cheapest of all these strategies is necessarily the globally optimal strategy $T^{opt}$. Thus, if there are $n$ components we can compute $T^{opt}$ in $O((n \log n)2^n)$. This, of course, is practical only when $n$ is small. However, our intent is to use this result for computing optimal strategies for hierarchical systems. As we shall see below, in that context, $n$ is indeed small.

### 5.2 THE CONDITIONAL EXPECTED COST OF REPAIR

Note that the expected cost $EC$ in Equation 10 is the overall expected cost. This cost is computed assuming that no observations have been made of the system as yet. In particular, it is not yet known whether the system is faulty.

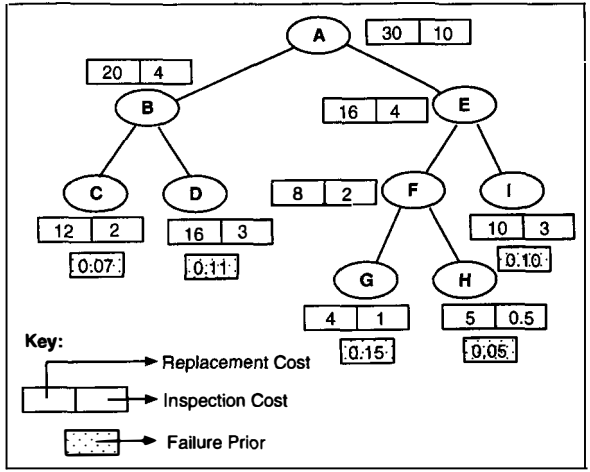

Figure 1: A hierarchical system model.

Consider instead the expected cost given that that we know the system is faulty. This expected cost estimate will be needed later when computing optimal hierarchical repair strategies. For any strategy $T$, let $EC^f(T)$ denote the expected cost of repair given that we know the system is faulty. Note that the observation "System is faulty" is exactly $S_0 = \langle X_0 = b \rangle$ (see Section 2). We see that $EC(T)$ and $EC^f(T)$ are related as follows:

$$EC(T) = P(X_0 = b) \times EC^f(T) + P(X_0 = ok) \times 0$$

Hence:

$$\begin{aligned} EC^f(T) &= \frac{EC(T)}{P(X_0 = b)} \quad (11) \\ &= \frac{EC(T)}{P(S_0)} \\ &= \frac{EC(T)}{1 - P(M_{[1,n]} = ok)} \end{aligned}$$

Note that for any strategy $T$, $EC^f(T)$ and $EC(T)$ are related by the constant $\frac{1}{1-P(M_{[1,n]}=ok)}$. Thus, the strategy $T^{opt}$ with the lowest possible value of $EC$ is also the strategy with the lowest value of $EC^f$.

## 6 HIERARCHY

Hierarchical systems are widely used in engineering practice. Hierarchies serve two purposes – they make modeling easier by separating the system into composable subsystems. In addition, an algorithm that operates on a system model to compute some property of interest can often be modified to exploit the hierarchy to give computational gains. We will first present a generalization of our system repair problem for hierarchical systems. We will then develop a hierarchical version of the optimal repair strategy algorithm. The hierarchy is exploited to make the algorithm tractable.

We begin by defining a hierarchical model. A *hierarchical component model* consists either of an *atomic model* or a *subcomponent model*.

If the component $\mathcal{C}$ is modeled atomically, we specify a probability of failure of the component $p$, a cost of replacement $c$ and a cost of inspection $d$.

If the component $\mathcal{C}$ is modeled as consisting of subcomponents, we do the following:

1. Specify hierarchical models (recursively) for each of the subcomponents $\mathcal{C}_i^s$ of $\mathcal{C}$.
2. Specify a cost of replacement $c$ of $\mathcal{C}$ and an inspection cost $d$.

We assume that a component works normally iff all its subcomponents are working normally. In other words, a component is in the *ok* state iff all its subcomponents are in the *ok* state. If any of the subcomponents are in the broken state the component is assumed to be in the broken state. Note that the probability of failure of the component can easily be computed from the subcomponent probabilities. Also, note that the top level component in the hierarchy represents the entire system. A *hierarchical system model* is simply the hierarchical component model for this top level component. Figure 1 is an example of a hierarchical system model. The tree in the figure represents the hierarchy tree of the system. Each node represents a component. The replacement cost and inspection cost of each component are marked next to it. In addition, the prior probability of failure for each of the leaf level components is also specified. Note that the prior probability of failure of the non-leaf components in the hierarchy tree can be computed from the probabilities at the leaves.

We now define a *hierarchical repair plan*. A hierarchical repair plan for a component specifies an action that will repair a component if it has been observed and found to be broken. The action specified is either:

- Replacement of the entire component.

or

- A strategy for repair of the subcomponents. As we saw before, a strategy specifies an order in which to repair the subcomponents. In addition, it specifies whether each subcomponent is to be inspected before repair or not.

If a strategy specifies that a subcomponent is not to be inspected before repair, it is simply replaced. If a strategy specifies that the subcomponent is to be inspected then the inspection procedure of the subcomponent is carried out before it is repaired. If the result of the inspection is that the subcomponent is *ok* then the subcomponent needs no further attention.

If the result of the inspection is that the subcomponent is broken, then the subcomponent is repaired according to a hierarchical repair plan specified for the

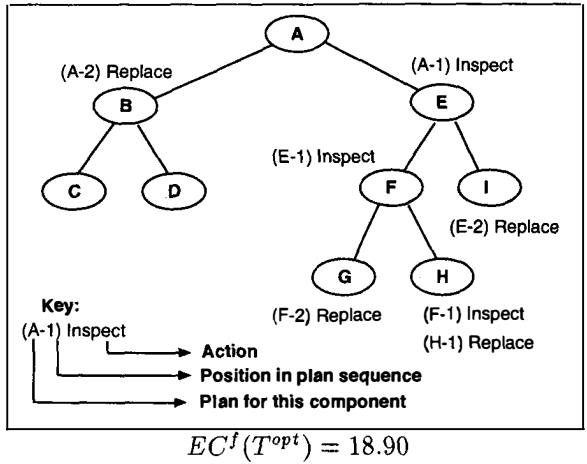

$$EC^f(T^{opt}) = 18.90$$

Figure 2: An (optimal) hierarchical system repair plan for system of Fig 1.

subcomponent. A hierarchical repair plan for a component thus includes the specification of a hierarchical repair plan for each of the subcomponents that are inspected by the plan.

Figure 2 is a possible hierarchical repair plan for the system. This plan specifies that if the system $A$ is known to be faulty we first repair $E$ after inspection and then, if $A$ is still faulty, replace $B$ without inspection.

The repair of $E$ proceeds as follows: If $E$ is found to be faulty after inspection we first repair $F$ after inspection and then, if $E$ is still faulty, replace $I$ without inspection. If $F$ is found to be faulty after inspection we first repair $H$ after inspection and then, if $F$ is still faulty, replace $G$ without inspection. If $H$ is found to be faulty after inspection it is replaced.

An optimal hierarchical repair plan for a component is the hierarchical repair plan with least expected cost. An *optimal hierarchical system repair plan* is simply the optimal hierarchical component repair plan for the top level component in the hierarchy tree. The repair plan shown in Figure 2 is also the optimal repair plan for the system.

### 6.1 COMPUTING THE OPTIMAL HIERARCHICAL PLAN

We will now describe a way of computing the optimal hierarchical repair plan for a component from the optimal hierarchical repair plans of its subcomponents. This procedure can then be used in a bottom-up traversal of the hierarchy tree to compute the optimal hierarchical system repair plan.

Say component $C$ has $k$ subcomponents $C_1^s, C_2^s, \ldots, C_k^s$. The replacement cost of the component is $c$. Say the optimal hierarchical plan for each subcomponent $C_i^s$ has already been computed and the cost of the plan is $H_i^s$.

We first compute an optimal strategy $T^{opt}$ in which to fix the subcomponents $C_i^s$. As we saw in Section 5.1, the optimal strategy $T^{opt}$ and its cost $EC(T^{opt})$ can be computed in $O((k \log k)2^k)$. The computation takes into account the optimal hierarchical repair cost $H_i^s$, the inspection cost $d_i^s$ and the replacement cost $c_i^s$ for each of the subcomponents $C_i^s$.

The cost estimate $EC(T^{opt})$ is the cost estimate for repairing $C$ given no evidence. Consider the situation where $C$ has been inspected and found to be broken. In this case, the cost estimate needs to be conditioned on this knowledge. As we saw in Section 5.2, the conditional cost estimate $EC^f(T^{opt})$ is given by:

$$\begin{aligned} EC^f(T^{opt}) &= \frac{EC(T^{opt})}{1 - P(M_{[1,n]} = ok)} \\ &= \frac{EC(T^{opt})}{1 - \Pi_{1 \leq i \leq n}(1 - p_i^s)} \end{aligned}$$

The optimal hierarchical plan specifies the optimal repair action (and accompanying cost) for a component given that it is broken. The two possible actions are: (a) replacement of the component and (b) repair of subcomponents. We can choose the better of the two options by simply comparing the replacement cost $c$ and the optimal cost $EC^f(T^{opt})$ of repairing subcomponents.

If $c \leq EC^f(T^{opt})$, then the optimal hierarchical repair plan for the component $C$ is to simply replace it if it has been found to be broken. The cost of the optimal hierarchical component repair plan in this case is $c$. If we find that $EC^f(T^{opt}) > c$ then the optimal hierarchical repair plan is to follow strategy $T^{opt}$. The cost of the hierarchical component repair plan for component $C$ in this case is $EC^f(T^{opt})$.

We note that we can compute the optimal hierarchical system repair plan by working up from the leaves of the hierarchy tree while computing the optimal component repair plan for each component. Say each component in the system can have at most $k$ subcomponents. Let us suppose the system has $n$ leaf level components in all. A tree with a branching factor of $k$ with $n$ leaf nodes has $O(n)$ nodes in the tree (including leaf nodes). So the complexity of computing the optimal hierarchical repair plan is $O(n(k \log k)2^k)$. Hence, for a fixed $k$, the optimal hierarchical repair plan can be computed in $O(n)$.

We have implemented the algorithm in Common Lisp. For testing purposes, we have also have implemented a random system generator that creates a system hierarchy with a user specified branching factor and a user specified tree depth. The repair costs, inspection costs and failure probabilities are chosen randomly from user specified intervals. The time taken to compute optimal repair strategies for systems of various sizes are shown in Table 1. The optimal policy for a system with a branching factor of 5 and a tree depth of 5 (i.e., with

Table 1: Running time of optimal hierarchical repair plan algorithm.

| $k$ | \\ | $d$ | |
|---|---|---|---|
| | 3 | 4 | 5 |
| 3 | 17 | 50 | 117 |
| 4 | 83 | 233 | 833 |
| 5 | 167 | 934 | 4750 |

$k$ : Branching Factor, $d$ : Tree Depth
Run time in milliseconds on a Sun 10/40.

3125 leaf level components) can be computed in about 5 seconds. Thus the algorithm scales well to systems with thousands of components.

## 7  DISCUSSION

In a general formulation of the repair problem, the pre-computation of an optimal repair strategy is intractable. The reason is that in a general formulation, there are no restrictions on the kind of system modeled. Since there is no special structure that we can take advantage of, we are reduced to considering each possible strategy in a combinatorial space of repair strategies to compute the optimal strategy.

There are two classes of approaches used to address this tractability problem. The first is to make some restricted formulation of the repair problem which still applicable in some domain of interest. The properties of the restricted formulation can then be exploited to develop tractable algorithms to compute repair strategies. The work in this paper falls into this class. In the second class of approaches, the diagnosis/repair problem is formulated as an interactive process. At each stage of the process, an action that is to be carried out immediately is chosen with a greedy heuristic or limited lookahead. The chosen action is then carried out and this leads to new information being obtained. This information is used to compute the next action to be carried out.

[Kalagnanam & Henrion, 1988] derive an optimality condition for the optimal repair strategy in a multi-component system which is assumed to have a single fault. The repair protocol is similar to the one described in this paper with the exception that only component replacements are allowed. There is no notion of inspection of components. Our work generalizes their result to the case of multiple independent failures. It also introduces a formulation of component inspection and extends the scope of the algorithm to hierarchical systems.

[Heckerman et al, 1995] formulate repair as an interactive process. The system is modeled with a Bayesian network and both component replacement and information gathering actions are possible. An action is chosen at each step of the process with a myopic heuristic. The heuristic computes the least cost action to take next assuming that the current fault in the system is a single fault. The restricted system behavior we have proposed in this paper corresponds to a restricted form of Bayesian network in their framework. When no component inspections[1] are allowed we have developed a polynomial time algorithm for computing the optimal strategy. This algorithm is thus a tractable solution to a special case of the problem attacked by [Heckerman et al, 1995].

The optimality result of Equation 6 is potentially applicable within their framework as an improved heuristic for choosing actions myopically. Instead of assuming a single fault the improved heuristic would allow for multiple faults.

The work in the model-based diagnosis community ([Hamscher et al, 1992]) has also addressed the repair problem as an interactive process. [deKleer & Williams, 1987] introduce an entropy based method for observation planning. [Friedrich & Nejdl, 1992] develop a set of greedy algorithms for choosing observation and repair actions in interactive model-based diagnosis. Their approach explicitly considers downtime costs in case of unanticipated failure. Hence their repair scheme implicitly includes a notion of preventive maintenance. [Poole & Provan, 1991] use repair actions to partition the world into a set of classes. All the worlds in a class result in the same action response. The diagnosis problem now becomes one of determining which class the current state of the system falls into. The action response can then be looked up. [Yuan, 1993] proposes a decision theoretic framework for modeling interactive model-based diagnosis. At each step of the diagnosis a decision model in the form of an influence diagram is synthesized and solved to compute the next action. The model is successively refined along the system hierarchy using a single fault assumption until the fault is located. Our formulation of repair in this paper is more restricted than in any of these model-based approaches.

[Srinivas & Horvitz, 1995] develop a formulation of model-based diagnosis in hierarchical systems and develop a linear time algorithm for pre-computation of an optimal repair strategy. A particular repair protocol is assumed: Repair begins when the system exhibits an anomalous output for some input. The repair process consists of successively repairing components of the system until the output is no longer anomalous for the same input. The hierarchy is exploited to gain tractability when computing the optimal-repair strategy. The algorithm is tractable if the branching factor of the system hierarchy is small. The hierarchical algorithm we have developed in this paper addresses a special case of this general formulation. The constant in the linear time algorithm for this special case is far smaller than the constant in the general formulation.

---
[1] Called component observations in their paper.

## 7.1 DEPENDENT FAULTS

This paper has concentrated on the situation where component failures are independent. Note however that the optimality condition (Equation 5) applies in the general case where the component failures may be dependent.

Consider the situation where a model for the dependencies between the failures of the components is available in the form of a Bayesian network $B$. Thus, $B$ is a model for $P(M_1, M_2, \ldots, M_n)$. The optimality condition of Equation 5 can be simplified to:

$$\begin{aligned} c_j[P(M_j = ok \mid R_{ok}) - \\ P(M_j = ok, M_{j+1} = ok \mid R_{ok})] \\ \leq \quad c_{j+1}[P(M_{j+1} = ok \mid R_{ok}) - \\ P(M_{j+1} = ok, M_j = ok \mid R_{ok})] \end{aligned} \quad (12)$$

Given a repair sequence $T$ we can check whether the condition holds at the position $j$ as follows. Declare evidence $R_{ok}$ (i.e., $M_{[j+2,n]} = ok$) in the network $B$. Do a network inference and look up the probabilities $P(M_j = ok \mid R_{ok})$ and $P(M_{j+1} = ok \mid R_{ok})$. Subsequently declare additional evidence $M_j = ok$ and do another network inference to compute $P(M_{j+1} = ok \mid M_j = ok, R_{ok})$. Note that the probability $P(M_j = ok, M_{j+1} = ok \mid R_{ok})$ can now be computed as:

$$P(M_j = ok, M_{j+1} = ok \mid R_{ok}) = \\ P(M_{j+1} = ok \mid M_j = ok, R_{ok})P(M_j = ok \mid R_{ok})$$

We now have the quantities required to check whether the condition holds. Thus the verification of the optimality condition at any point in the sequence $T$ can be accomplished with 2 network inferences.

If the optimality condition does not hold at position $j$, flipping the position of $C_j$ and $C_{j+1}$ will lead to a better repair sequence. If we consider doing this repeatedly till quiescence is reached then the resulting sequence will be a local optimum. A good starting sequence might the sequence $T^{ind}$ computed assuming that the component failures are independent. This can be done by initially doing network inference with no evidence in the network. That gives us the priors $p_i = P(M_i = b)$ for every node in the network and thus the sequence sorted by increasing order of $c_i \frac{1-p_i}{p_i}$ can be computed.

However, certain questions still need to be addressed. Firstly, it is not clear that the sequence produced at quiescence is necessarily the global optimum. The second question is whether the number of network propagations required is tractable in the worst case.

With regard to the second question: the first naive estimate is that the number of propagations is $O(n!)$ since there are only $n!$ sequences. However, we can observe from the structure of Equation 3 that if $k$ components have been fixed then the optimal repair sequence of the remaining $n - k$ components does not depend on the order in which the first $k$ components were fixed. This allows dynamic programming to be used to construct a scheme which will compute the optimal strategy with $O(n2^n)$ network propagations. This is still exponential.

Our speculation is that we cannot do better without more structure (for eg, specific network topologies) in the problem. A promising direction seems to be to adapt an exact algorithm for computation of the optimal strategy in the case of dependent faults to have limited lookahead and anytime characteristics. We plan to look into these topics in the future.